\documentclass[conference]{IEEEtran}
\IEEEoverridecommandlockouts

\usepackage{cite}
\usepackage{amsmath,amssymb,amsfonts}
\usepackage{wrapfig}
\usepackage{graphicx}
\usepackage{textcomp}
\usepackage{xcolor}
\usepackage{url}
\usepackage{tikz} 
\usepackage{todonotes}                  
\usepackage{caption}
\usepackage{algorithmicx}
\usepackage{algpseudocode} 

\usepackage{subcaption}                 
\usepackage[caption=false,font=footnotesize]{subfig}
\usepackage{siunitx}   					

\usepackage{cleveref}
\crefname{figure}{fig.}{figs.}
\Crefname{figure}{Fig.}{Figs.}

\crefname{table}{tab.}{tabs.}
\Crefname{table}{Tab.}{Tabs.}

\crefname{equation}{eq.}{eqs.}
\Crefname{equation}{Eq.}{Eqs.}

\crefname{algorithm}{alg.}{algs.}
\Crefname{algorithm}{Alg.}{Algs.}

\begin{document}

\title{
\LARGE
Real-World Deployment of Massively Parallel Sampling-Based MPC\\for Contact-Rich Manipulation
}

\author{
Magnus Dierking$^{1,2}$,
Jo\~{a}o Carvalho$^{1,3}$,
An T. Le$^{1,6}$, 
Georgia Chalvatzaki$^{2,4,5}$ 
and 
Jan Peters$^{1,3,4,5}$
\thanks{
$^{1}$Intelligent Autonomous Systems Lab, TU Darmstadt, Germany;
$^{2}$Interactive Robot Perception \& Learning Lab, TU Darmstadt, Germany;
$^{3}$German Research Center for AI (DFKI); 
$^{4}$Hessian.AI;
$^{5}$Robotics Institute Germany (RIG).
$^{6}$College of Engineering and Computer Science, VinUniversity, Vietnam.
This work was funded by the German Federal Ministry of Education and Research Software Campus project
ROBOSTRUCT (16|S23067).
Corresponding author: Magnus Dierking, magnus.dierking@stud.tu-darmstadt.de.
}
}

\maketitle

\begin{abstract}
Sampling-based Model Predictive Control (SMPC) is a promising strategy for contact-rich robotic manipulation, combining gradient-free optimization with massively parallel GPU simulation. Yet, most prior work relies on simplified dynamics or remains confined to simulation.

We present an MPC framework that leverages JAX for large-scale parallelization and efficient computation, coupled with the high-fidelity MuJoCo MJX simulator, and deploy it on a Franka Research 3 executing the Push-T manipulation task through a complete real-to-sim-to-real pipeline. The MTP variant with structured global sampling outperforms unimodal baselines such as CEM, MPPI, and PS across tasks that require mode switching, both in simulation and on hardware. Furthermore, we evaluate online domain randomization within the MPC sample budget, showing that contact-initiation parameters yield interpretable adaptation signals, whereas global physics parameters provide feedback that is too weak for reliable exploitation at typical replanning frequencies. These findings highlight key challenges for sampling-based MPC in contact-rich manipulation—contact sensitivity, tight compute budgets, and the difficulty of obtaining informative domain-randomization signals in real time.

\end{abstract}


\section{Introduction}

Classical gradient-based MPC assumes smooth, differentiable dynamics—an assumption that breaks down in contact-rich manipulation, where dynamics are hybrid and nonsmooth. Sampling-based MPC (SMPC) circumvents this limitation by drawing candidate control trajectories, evaluating them via forward simulation, and updating the control distribution based on outcomes. Combined with JAX~\cite{jax} and GPU-accelerated physics engines such as MuJoCo~MJX~\cite{mjx-computation}, we can roll out thousands of trajectories in parallel, making real-time MPC feasible for robotics.

Despite growing interest, real-robot SMPC has focused primarily on locomotion. DIAL-MPC~\cite{xueFullOrderSamplingBasedMPC2024} and Whole-Body MPPI~\cite{alvarez2025real} achieve impressive results on quadruped jumping and climbing, while~\cite{feedback_mppi} extends the approach to drone flight. For manipulation, Pezzato et al.~\cite{pezzatoSamplingbasedModelPredictive2025} present the most relevant real-world study. However, their demonstrations on a dexterous anthropomorphic hand remain proofs of concept—the authors report limited GPU throughput and real-time control constraints that restricted trajectory smoothness and prevented effective parallelization.

Standard SMPC algorithms such as MPPI~\cite{williamsModelPredictivePath2017} and CEM~\cite{pinneriSampleefficientCrossEntropyMethod2020} rely on unimodal Gaussian proposals and often become trapped in local minima when the cost landscape is multimodal—a common occurrence in contact-rich manipulation. Model Tensor Planning (MTP)~\cite{leModelTensorPlanning2025} addresses these problems by augmenting local Gaussian sampling with graph-based global candidates, enabling a better exploration vs. exploitation tradeoff.
However, MTP has so far been evaluated only in simulation, leaving open questions about real-world applicability, scalability under real-time constraints, and sim-to-real gap sensitivity.
Contact-rich manipulation tasks such as Push-T expose these limitations directly, providing reproducible benchmarks in both simulated and physical settings.

This paper makes the following contributions:
\begin{itemize}
\item Simulation baseline comparison of MTP, MPPI, CEM, and Predictive Sampling (PS)~\cite{howellPredictiveSamplingRealtime2022} on tasks requiring multimodal exploration;
\item Real-robot deployment of SMPC for contact-rich manipulation using a high-fidelity physics backend (MuJoCo MJX) on the Push-T task with a Franka Research~3.
\item An empirical analysis of online domain randomization within SMPC, identifying which physics parameters yield interpretable adaptation signals and why learning from them remains difficult under contact-rich dynamics and tight compute constraints.
\end{itemize}
\section{Background}
\label{sec:background}

\subsection{Sampling-Based MPC}

We consider a discrete-time system $\mathbf{x}_{t+1} = \mathbf{g}(\mathbf{x}_t,
\mathbf{v}_t)$, with state $\mathbf{x}_t$ and control input $\mathbf{v}_t \sim
\mathcal{N}(\mathbf{u}_t, \boldsymbol{\Sigma})$.
Sampling-based MPC seeks a mean sequence $\mathbf{U}_k=[\mathbf{u}_0, \ldots, \mathbf{u}_{T-1}]$ that minimises an expected
trajectory cost $S(\mathbf{V}_k)$ over horizon $T$ by evaluating $N$ candidate
trajectories.

The Information-Theoretic MPC (IT-MPC) framework~\cite{williamsInformationTheoreticMPC2017}
grounds this in the Gibbs Variational Principle,
\begin{align}
  -\lambda\,\mathcal{F}(S,p,\lambda)
  \;\leq\;
  \mathbb{E}_{\mathbb{Q}}[S(\mathbf{V})] + \lambda\,\mathrm{KL}(\mathbb{Q}\|\mathbb{P}),
\end{align}
where $\mathcal{F}$ is the free energy of cost $S$ under prior $\mathbb{P}$ and
$\lambda{>}0$ is a temperature.
Minimising the right-hand side over Gaussian proposals yields exponential
reweighting
\begin{align}
  \mathbf{U}_k \;\leftarrow\; \sum_{i=1}^N w^i \mathbf{V}^i_k,
  \quad
  w^i \propto \exp\!\bigl(-S(\mathbf{V}^i_k)/\lambda\bigr),
  \label{eq:itmp-update}
\end{align}
which recovers MPPI~\cite{williamsModelPredictivePath2017} under Gaussian
perturbations.
CEM~\cite{pinneriSampleefficientCrossEntropyMethod2020} replaces soft reweighting
with hard elite selection (top-$E$ trajectories); PS~\cite{howellPredictiveSamplingRealtime2022} sets the next mean to the
lowest cost rollout.

\subsection{Model Tensor Planning}

Using a Gaussian distribution to model control sequences can lead to local minima.
MTP~\cite{leModelTensorPlanning2025} addresses this issue by mixing local
Gaussian samples with global \emph{tensor-path} samples $\mathbf{V}^G$ drawn from a layered
graph over the control space $\mathcal{U}$.
These are combined with local perturbations $\mathbf{V}^L \sim
\mathcal{N}(\mathbf{U}, \alpha\boldsymbol{\Sigma})$ into a joint batch of control signals
$\mathbf{V} = [\mathbf{V}^G;\,\mathbf{V}^L]$.
The update applies CEM-style elite selection followed by MPPI reweighting over
elites, and the best sample is applied (as in PS).
A mixing coefficient $\beta \in [0,1]$ controls the global/local budget split.
Near a low-cost mode, local samples dominate the elite set and the algorithm behaves
like MPPI; near a local minimum, global samples enter the elites and shift the
mean across cost modes. 

\subsection{Domain Randomization in Sampling-Based MPC}

Domain randomization improves sim-to-real robustness by considering a
distribution of physics parameters $h(\boldsymbol{\xi})$.
Abraham et al.~\cite{abrahamModelBasedGeneralizationParameter2020} show that we can extend IT-MPC as
\begin{align}
  -\lambda\,\mathcal{F}(S,p,\lambda)
  \;\leq\;
  \mathbb{E}_{h(\boldsymbol{\xi})}\!\Big[
    \mathbb{E}_{\mathbb{Q}}[S(\mathbf{V})] + \lambda\,\mathrm{KL}(\mathbb{Q}\|\mathbb{P})
  \Big],
\end{align}
promoting control sequences that are robust across the domain distribution.
In practice, this amounts to running parallel rollouts over $D$ randomised
environments and aggregating per-domain costs before the weight update~\eqref{eq:itmp-update}.

\section{Framework \& Methodology}
\label{sec:method}

\begin{figure}[t]
    \centering

    \begin{subfigure}[b]{0.99\linewidth}
        \centering
        \begin{tikzpicture}
        \node[draw=none, rounded corners=10pt, clip, inner sep=0pt]
          {\includegraphics[width=0.5\textwidth]{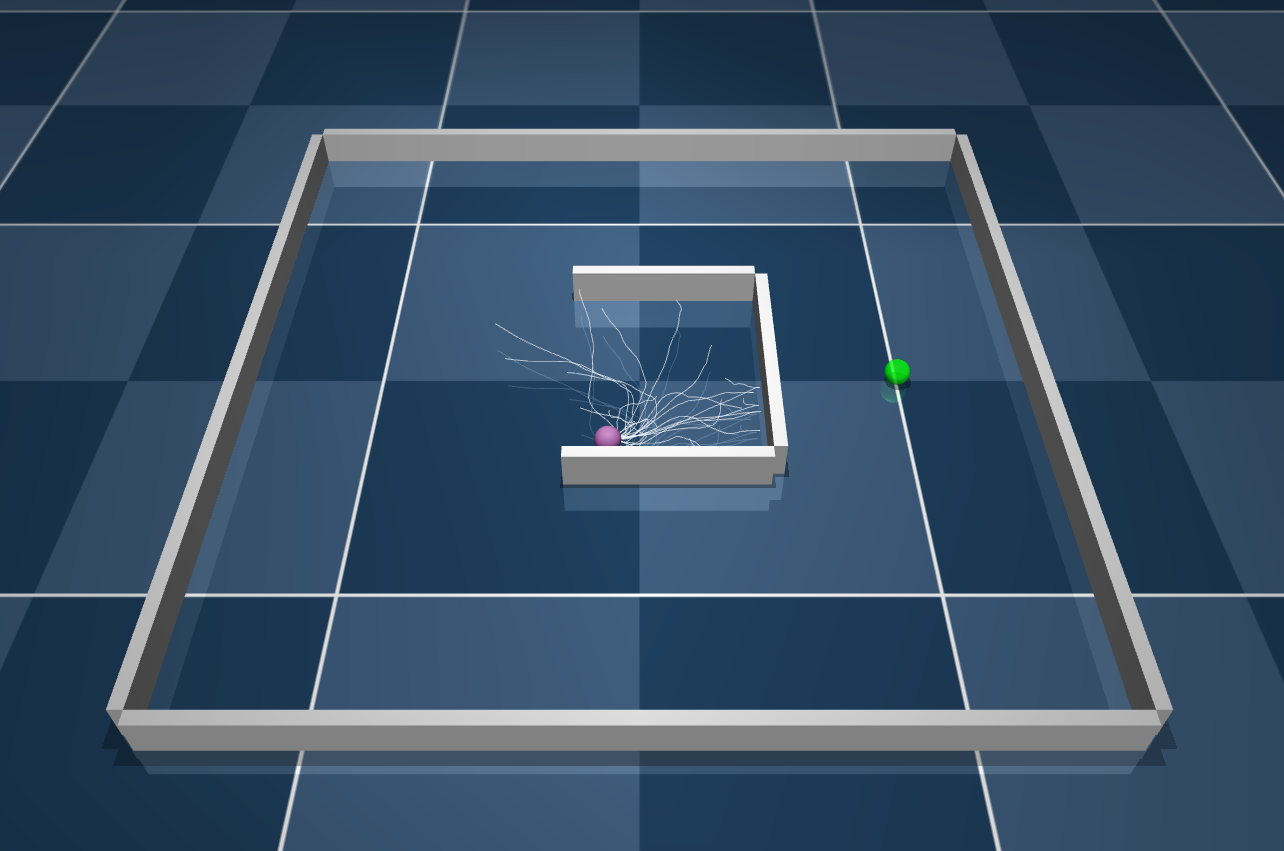}};
        \end{tikzpicture}
        \caption{Bugtrap Escape. Green sphere is the goal. Grey are trajectory samples.}
        \label{fig:bugtrap-intro}
    \end{subfigure}
    \\
    \vspace{-.2cm}
    \begin{subfigure}[t]{0.48\linewidth}
        \vspace{0pt}
        \centering
        \begin{tikzpicture}
        \node[draw=none, rounded corners=10pt, clip, inner sep=0pt]
          {\includegraphics[width=\textwidth,height=4cm,keepaspectratio]{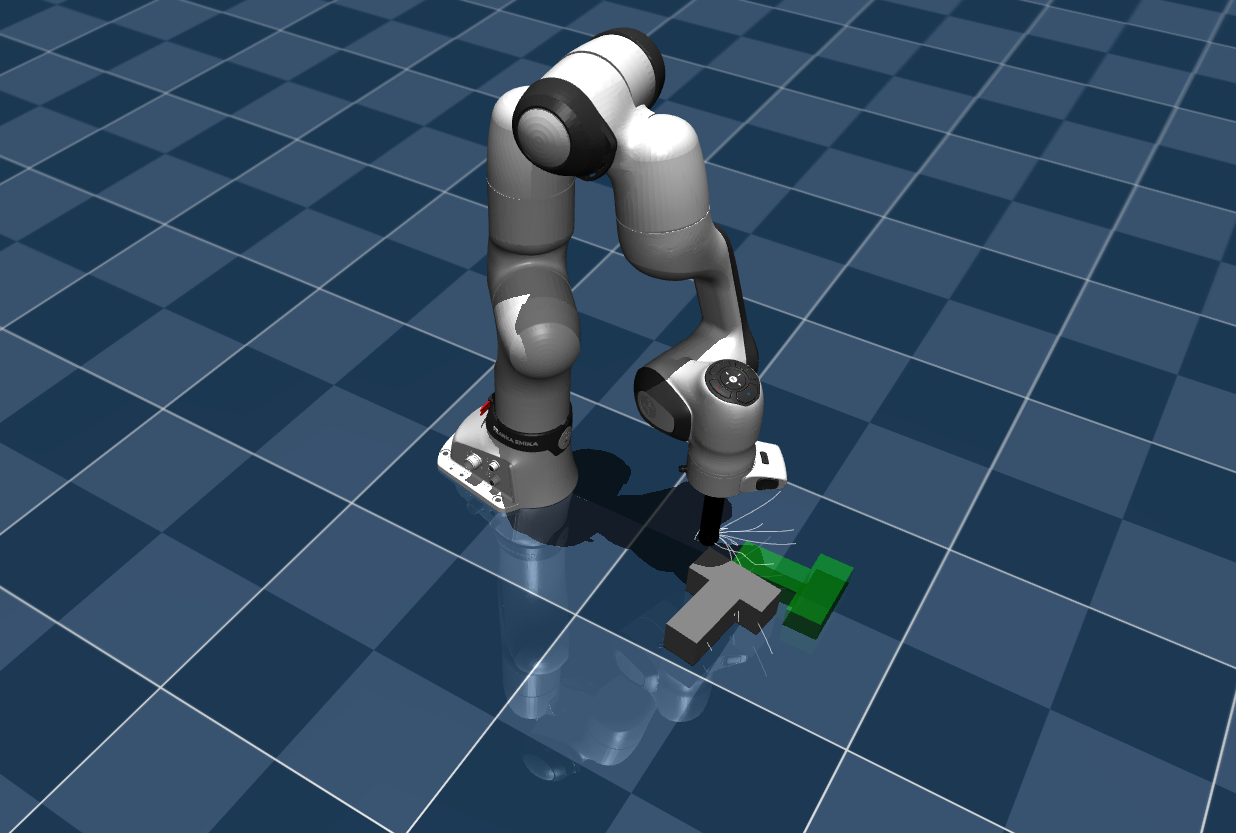}};
        \end{tikzpicture}
        \caption{Push-T in simulation. The green object is the goal pose.}
        \label{fig:pusht-intro}
    \end{subfigure}
    \hfill
    \begin{subfigure}[t]{0.48\linewidth}
        \vspace{0pt}
        \centering
        \begin{tikzpicture}
            \node[draw=none, rounded corners=10pt, clip, inner sep=0pt]
              {\includegraphics[trim={0 10cm 0 20cm},clip,width=\textwidth,height=3.0cm,keepaspectratio]{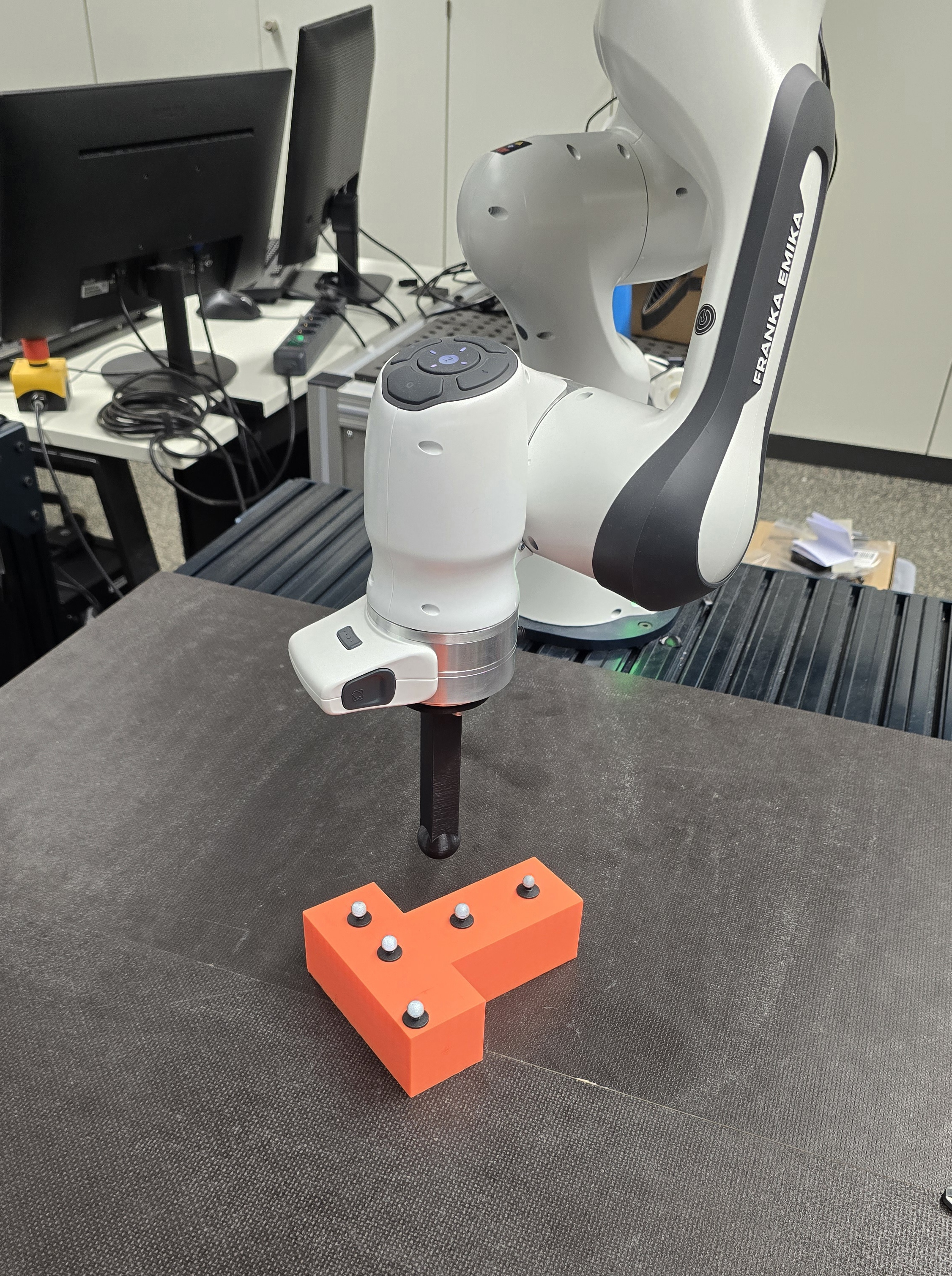}};
        \end{tikzpicture}
        \caption{Push-T real-world setup, with a FR3 arm, custom end-effector, and Mocap markers.}
        \label{fig:fr3-sketch}
    \end{subfigure}

    \caption{Simulation environments and real-world setup.}
    \label{fig:sim-envs}
\end{figure}

\subsection{Framework and Environments}

Our framework is built on top of Hydrax~\cite{hydrax}, a JAX-based SMPC framework using MuJoCo
MJX~\cite{mjx-computation} as the dynamics backend, enabling end-to-end JIT
compilation and \texttt{vmap}-vectorized rollouts across thousands of parallel
trajectories on GPU without hand-crafted dynamics models. We contribute two environments, framework optimizations, and a complete real-robot pipeline.

\textbf{Bugtrap Escape} (\cref{fig:bugtrap-intro}) consists of a 2D point mass inside a U-shaped trap built
from physical box primitives. The cost penalizes external contact forces and
distance to goal; starting inside the trap, the agent must discover an escape
route—a direct test of mode-switching capability.


\textbf{Push-T} (\cref{fig:pusht-intro}) requires pushing a rigid T-shaped block to a target pose using
a Franka Research~3 arm (\cref{fig:fr3-sketch}).
We use a $7$\,DoF free-floating rigid body (with orientation as a quaternion) 
that captures realistic friction, mass distribution, and full table-contact
dynamics. The cost combines T-pose alignment, end-effector proximity to the
object, and a retraction term to prevent singularities. We sample end-effector
twists and map them to joint velocities with differentiable inverse kinematics.


\subsection{Real-to-Sim-to-Real Pipeline}

\begin{figure}[t]
    \centering
    \includegraphics[trim={0  0cm 0 0cm},width=0.99\linewidth]{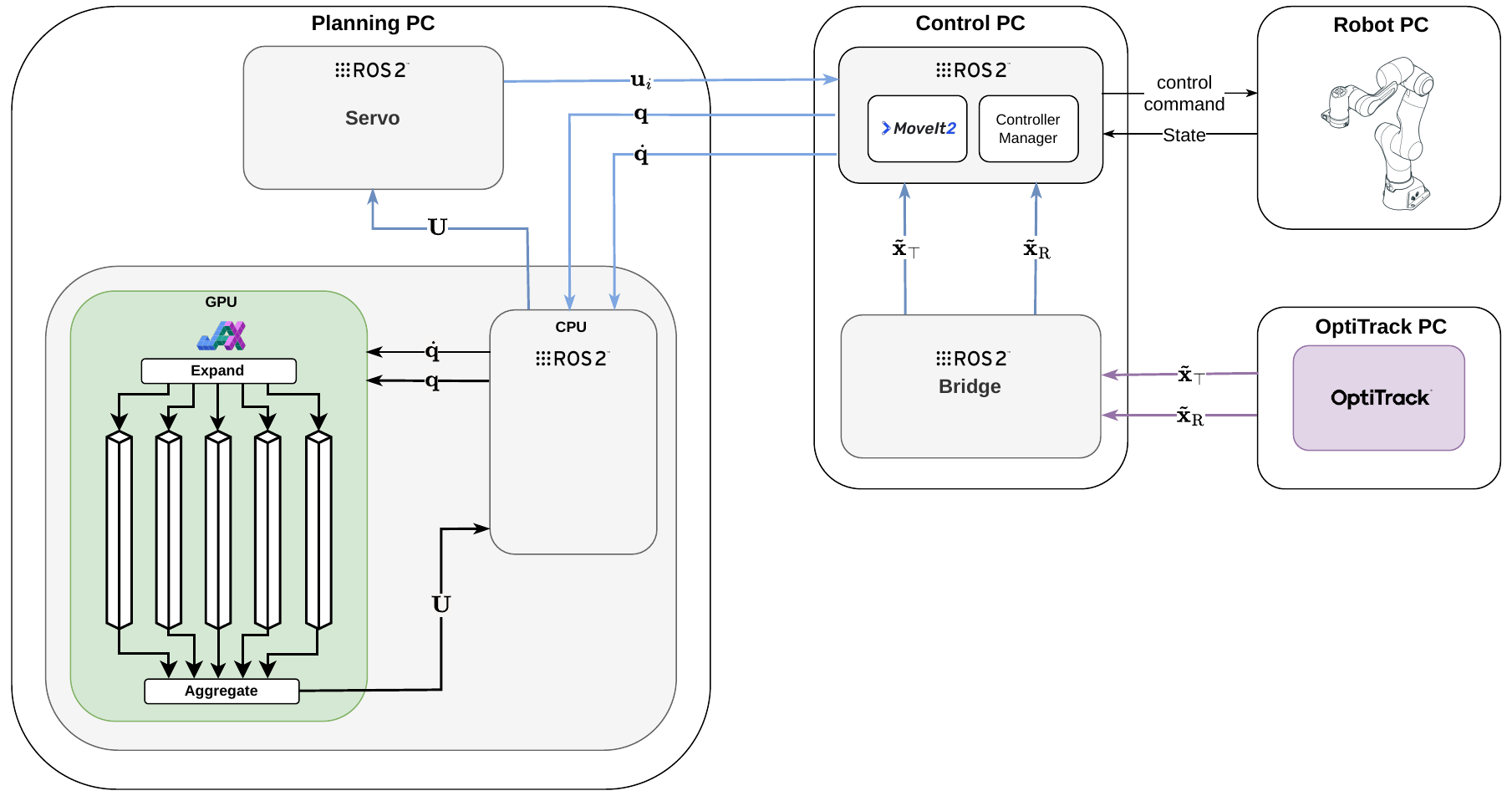}
    \caption{Real-world MPC experimental system overview based on ROS2. Blue arrows indicate ROS topic-based communication between processes. Purple arrows indicate Mocap streaming.}
    \label{fig:pushT-system}
    \vspace{-0.5cm}
\end{figure}

The real-robot system couples the planning loop to a Franka Research~3 and a 3D printed T through
four asynchronous ROS~2~\cite{ros2} processes (\cref{fig:pushT-system}).
Two \emph{control nodes} run MoveIt2~\cite{moveit} Servo for
$1\,\si{\kilo\hertz}$ Cartesian velocity streaming and select the current
horizon step at $50\,\si{\hertz}$. A \emph{bridge node} publishes Mocap poses (via OptiTrack) at $100\,\si{\hertz}$.
The \emph{planning node} executes the JIT-compiled MPC on an NVIDIA RTX 5090 GPU at $8$--$10\,\si{\hertz}$.

Because planning and servo nodes are decoupled, the servo node interpolates
within the published action trajectory based on elapsed time, maintaining
smooth control between replans. Hand--eye calibration~\cite{34770}
aligns the OptiTrack world frame with the
robot base frame.

\subsection{Domain Randomization}

We implement a structured domain randomization (DR) interface supporting per-body, per-geom, and
per-joint randomization. For parameters with derived quantities (mass,
inertia), we compile a separate MuJoCo model per domain tuple and extract the
resulting arrays into the batched pytree, ensuring physical consistency. Each
planning step rolls out $N$ trajectories over $D$ randomized domains
simultaneously; per-domain costs are aggregated via a weighted average, splitting the sample budget between trajectory samples and domains. 

Per-domain weights $w_d $ are either kept uniform or updated online based on the discrepancy between expected and observed system state. Concretely, we store the predicted object pose $\hat{\mathbf{T}}^{(d)}$ under each domain for the elite sample, and compare it against the observation $\mathbf{T}_\mathrm{obs}$ at the next replanning step:
\begin{align}
  w_d \propto \exp\!\bigl(-d_{\mathrm{SE(3)}}(\hat{\mathbf{T}}^{(d)}, \mathbf{T}_\mathrm{obs}) / \sigma^2\bigr),
\end{align}
where $d_{\mathrm{SE(3)}}$ is the geodesic pose error, and $\sigma$ a scaling parameter.

\section{Experiments}
\label{sec:experiments}

Our experiments address the following research questions: 
(RQ1) Does MTP outperform baselines that rely only on unimodal Gaussians on tasks that require switching between exploration and exploitation?
(RQ2) How do results transfer to real hardware under real-time constraints?
(RQ3) Which physics parameters yield informative domain-randomization feedback signals?

\begin{figure}[t]
    \centering
    \begin{subfigure}{0.3\linewidth}
        \centering
        \includegraphics[trim={40pt 40pt 40pt 40pt}, clip, width=\linewidth]{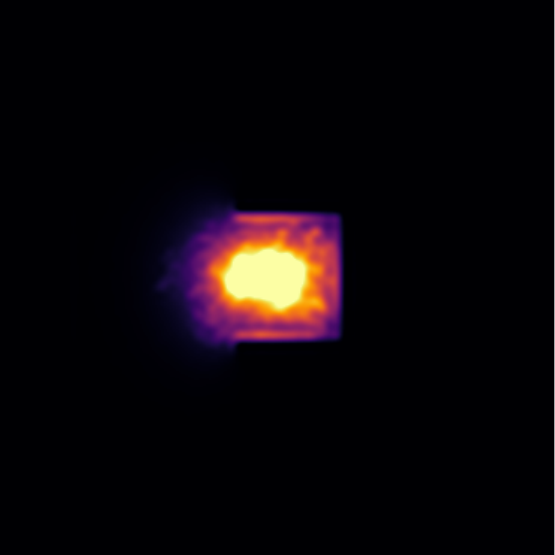}
        \caption{MPPI}
        \label{fig:bugtrap-mppi}
    \end{subfigure}
    \hspace{.6cm}
    \begin{subfigure}{0.3\linewidth}
        \centering
        \includegraphics[trim={40pt 40pt 40pt 40pt}, clip, width=\linewidth]{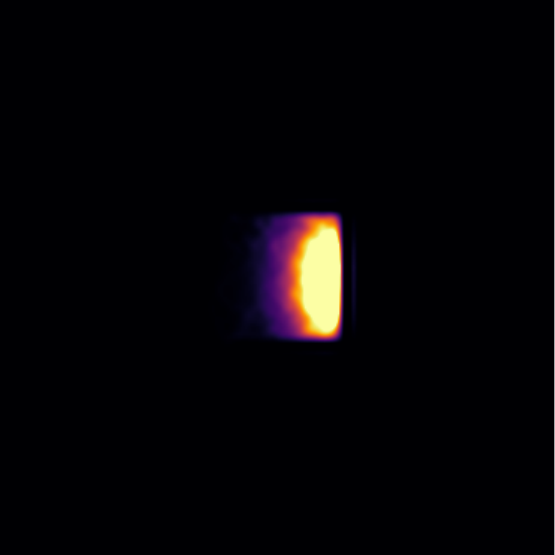}
        \caption{CEM}
        \label{fig:bugtrap-cem}
    \end{subfigure}
    \vspace{.3cm}
    \begin{subfigure}{0.3\linewidth}
        \centering
        \includegraphics[trim={40pt 40pt 40pt 40pt}, clip, width=\linewidth]{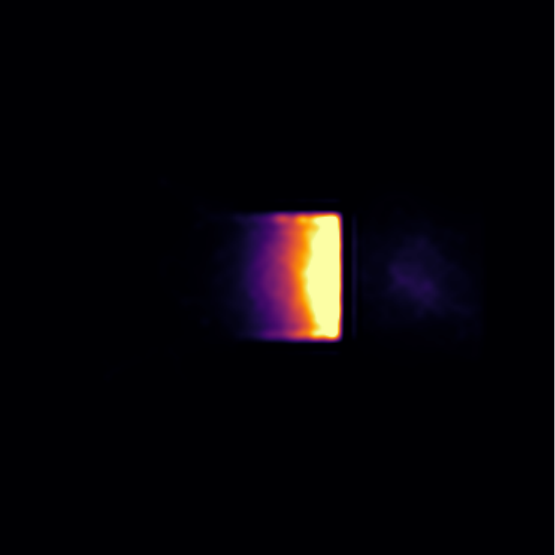}
        \caption{PS}
        \label{fig:escape-ps}
    \end{subfigure}
    \hspace{.6cm}
    \begin{subfigure}{0.3\linewidth}
        \centering
        \includegraphics[trim={40pt 40pt 40pt 40pt}, clip, width=\linewidth]{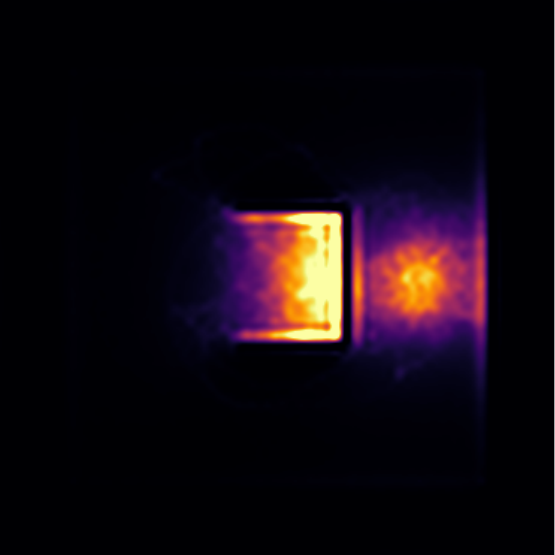}
        \caption{MTP}
        \label{fig:escape-mtp}
    \end{subfigure}
    \vspace{-0.4cm}
    \caption{Empirical state visitation distributions for the Bugtrap Escape task.}
    \label{fig:bug-trap-comparison}
    \vspace{-0.6cm}
\end{figure}

\subsection{Bugtrap Escape -- Simulation Experiment}

All SMPC algorithms run at $25~\si{\hertz}$ over a $0.72~\si{\sec}$ horizon for a total of $10{,}000$ planning steps.
We run $128$ samples across $6$ seeds with noise on the initial particle position; control sequences are sampled in velocity space.
The algorithm's behaviors can be seen by inspecting the state (position) visitations in \Cref{fig:bug-trap-comparison}.
MPPI and CEM fail in every seed: CEM's greedy elite selection traps it near the inner wall in a contact-avoiding limit cycle, while MPPI's full-sample averaging drives the particle toward the trap center where wall-collision costs and the goal attractor equilibrate.
PS escapes once but otherwise mirrors CEM.
Due to its high exploration capability, MTP escapes and reaches the goal in all seeds.

\begin{figure}[t]
    \centering
    \includegraphics[width=\linewidth]{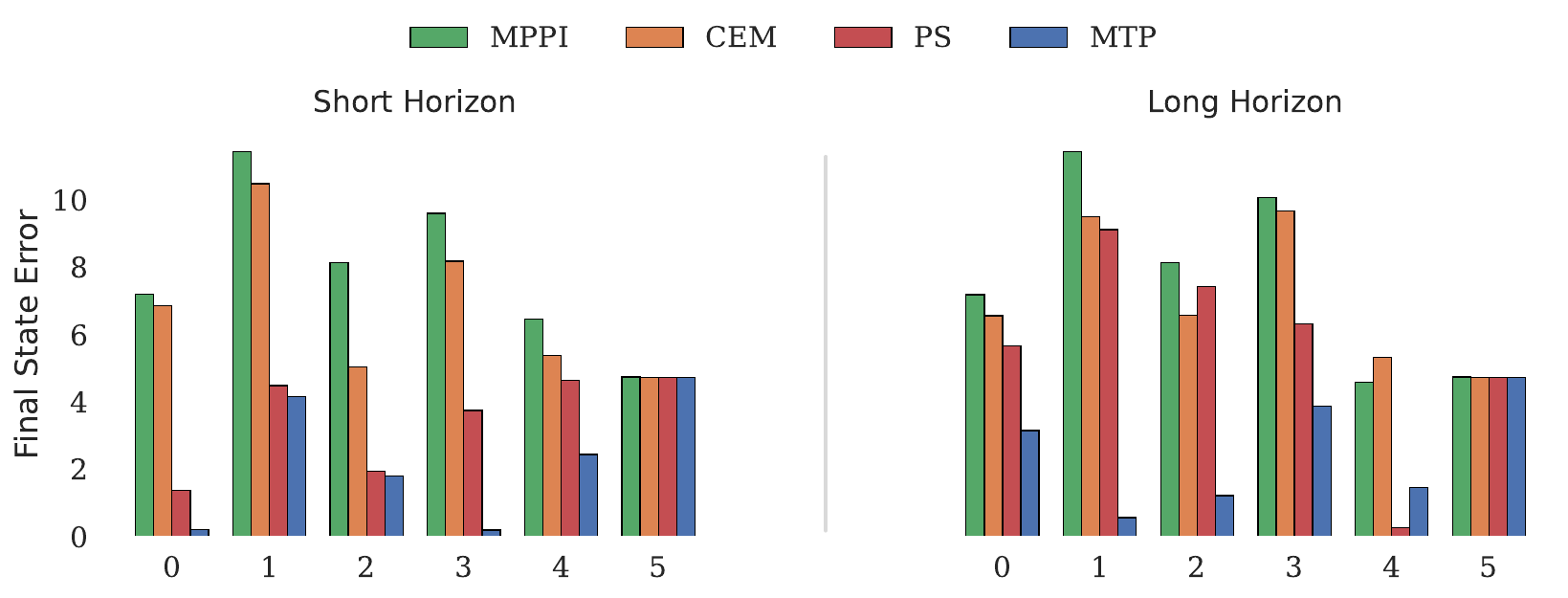}
    \caption{Sim-to-sim results for Push-T with long and short horizons. The number in the horizontal axis corresponds to a different initial state of the system.}
    \label{fig:results-sim-sim}
    \vspace{-0.5cm}
\end{figure}

\subsection{Push-T -- Simulation and Real-World Experiments}

\begin{figure*}[t]
    \centering
    \begin{subfigure}{0.25\linewidth}
        \centering
        \includegraphics[trim={0  0cm 0 6cm},width=\linewidth]{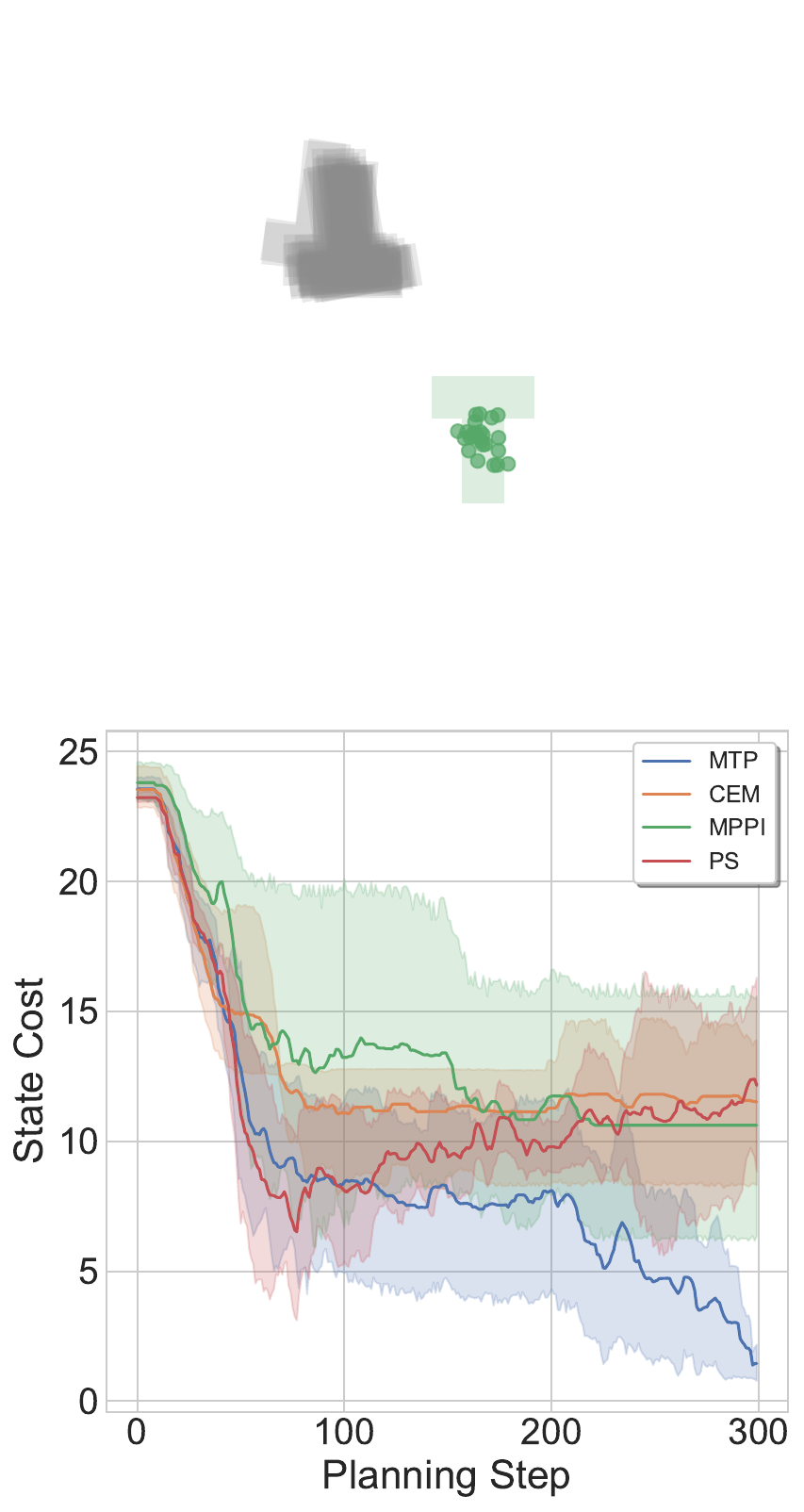}
    \end{subfigure}
    \hspace{.2cm}
    \begin{subfigure}{0.25\linewidth}
        \centering
        \includegraphics[trim={0  0cm 0 6cm},width=\linewidth]{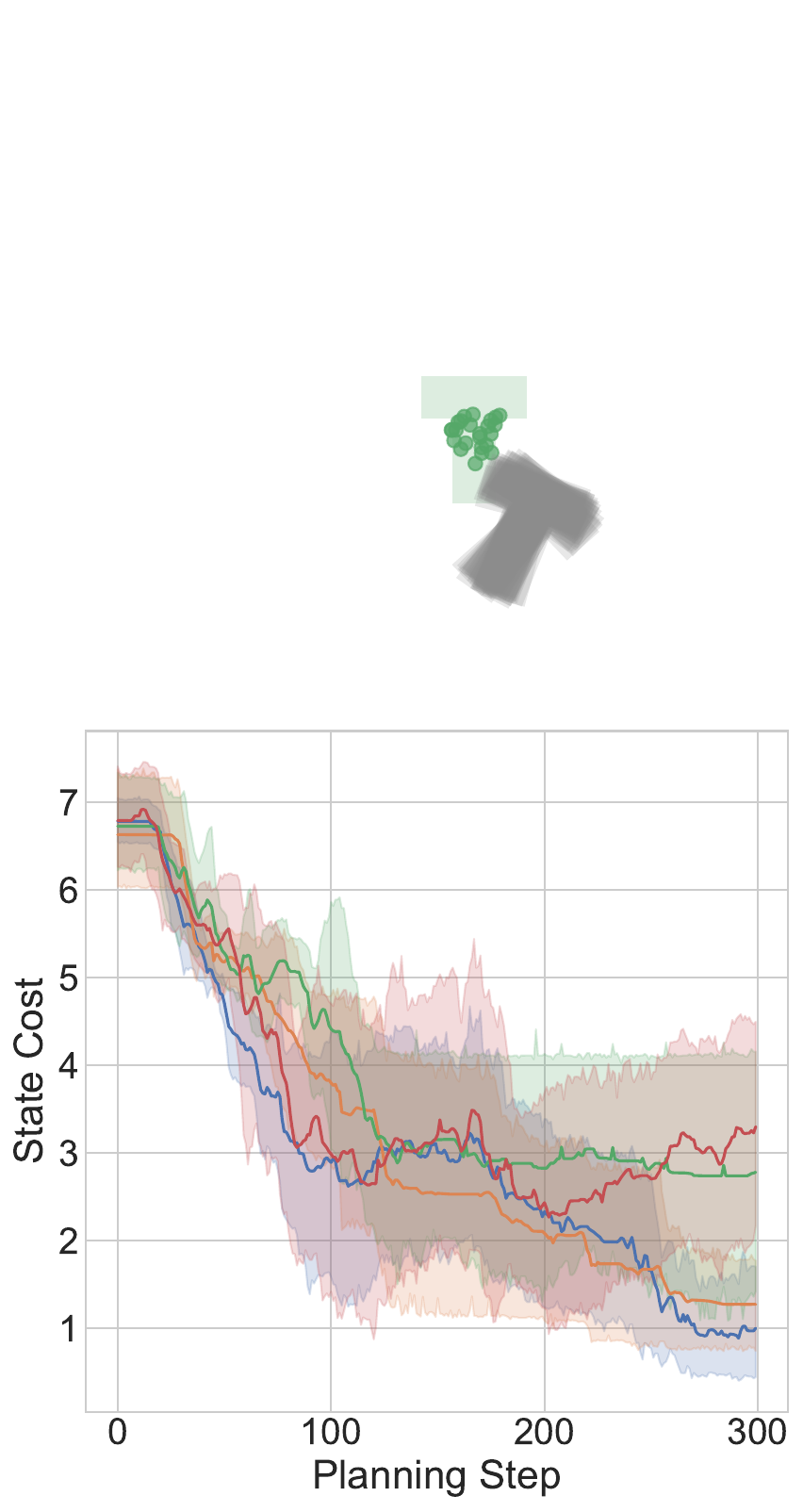}
    \end{subfigure}
    \hspace{.2cm}
    \begin{subfigure}{0.25\linewidth}
        \centering
        \includegraphics[trim={0  0cm 0 6cm},width=\linewidth]{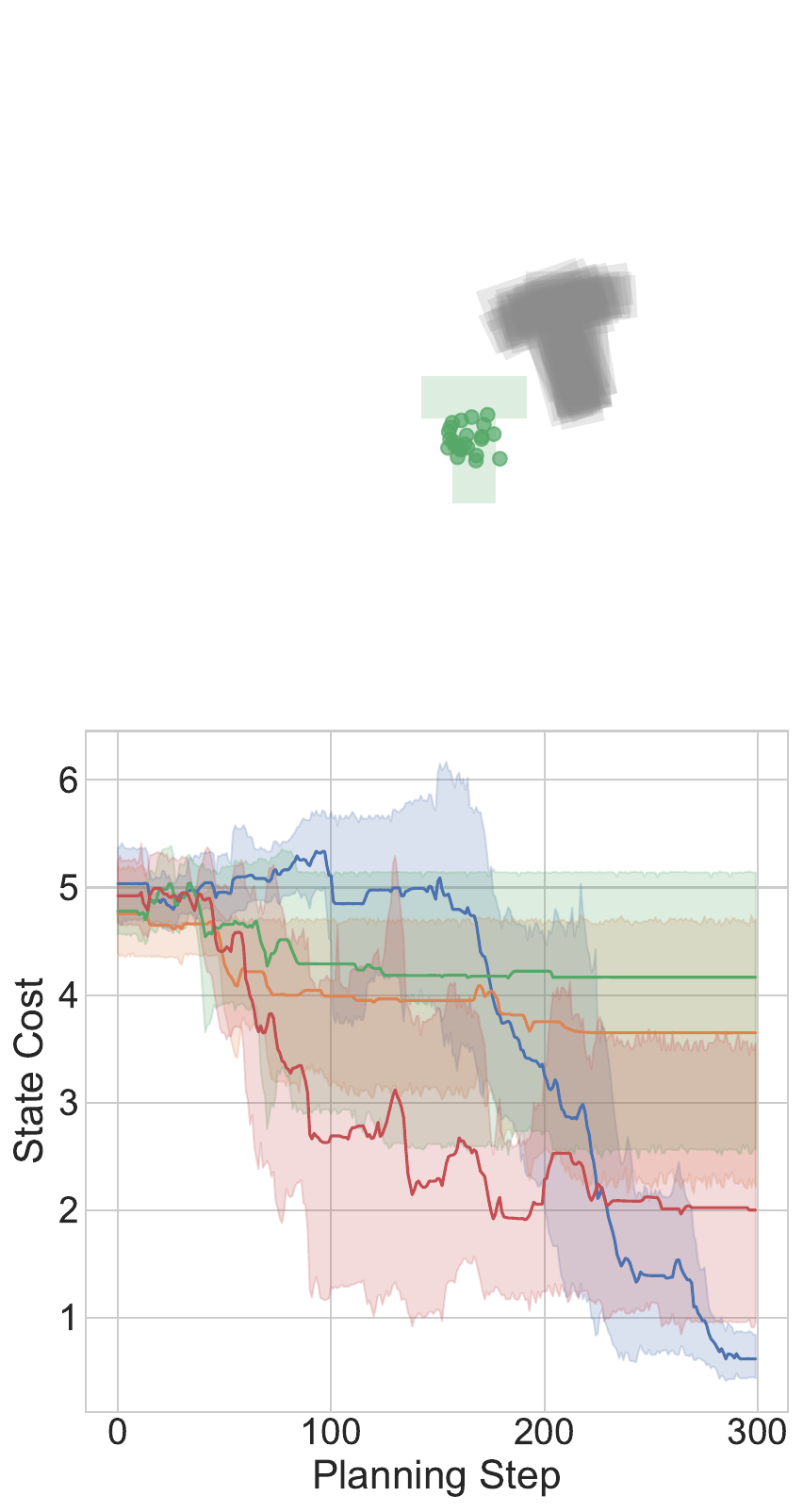}
    \end{subfigure}
    \caption{
    Real-to-sim-to-real results on the Push T task.}
    \label{fig:results-6dof}
    \vspace{-0.5cm}
\end{figure*}

In simulation, all methods use $256$ control samples and plan at $20~\si{Hz}$ with long ($0.625~\si{\sec}$) and short ($0.375~\si{\sec}$) horizons over $100$ planning steps.
\Cref{fig:results-sim-sim} presents the results.
To test the exploration capabilities, we deliberately set the end-effector and T's initial pose to hard-to-solve configurations.
MTP achieves the lowest final pose error across seeds that require mode switching, while unimodal baselines stall in local minima or degrade at shorter horizons.

In real-robot evaluations, we run $6$ seeds with small variations in T's initial pose and end-effector position.
We use $1024$ control sequence samples, planning at $8$--$10\,\si{Hz}$.
The MJX physics solver dominates runtime, requiring coarser timesteps 
and shorter horizons to meet replanning targets.
Integration steps above
$0.025\,\si{s}$ consistently lead to instability via deep single-step
penetrations that activate constraint-resolution accelerations rather than
physical friction. This forces a fundamental tradeoff: contact fidelity
requires small steps, and real-time replanning requires shorter horizons,
and both degrade receding-horizon performance if tasks need longer horizons for planning.

The results for different initial sampling distributions are reported in \Cref{fig:results-6dof}.
MTP achieves the lowest average final pose error and variance. MPPI frequently avoids contact entirely. PS
initiates contact aggressively but degrades from overshooting and brittle
contact exploitation. CEM produces the smoothest motions but stalls in local
minima after initial pose correction. 
 
\subsection{Domain Randomization Analysis}

\begin{figure}[t]
  \centering
  \includegraphics[trim={0pt 0pt 0pt 20pt}, clip, width=0.6\linewidth]{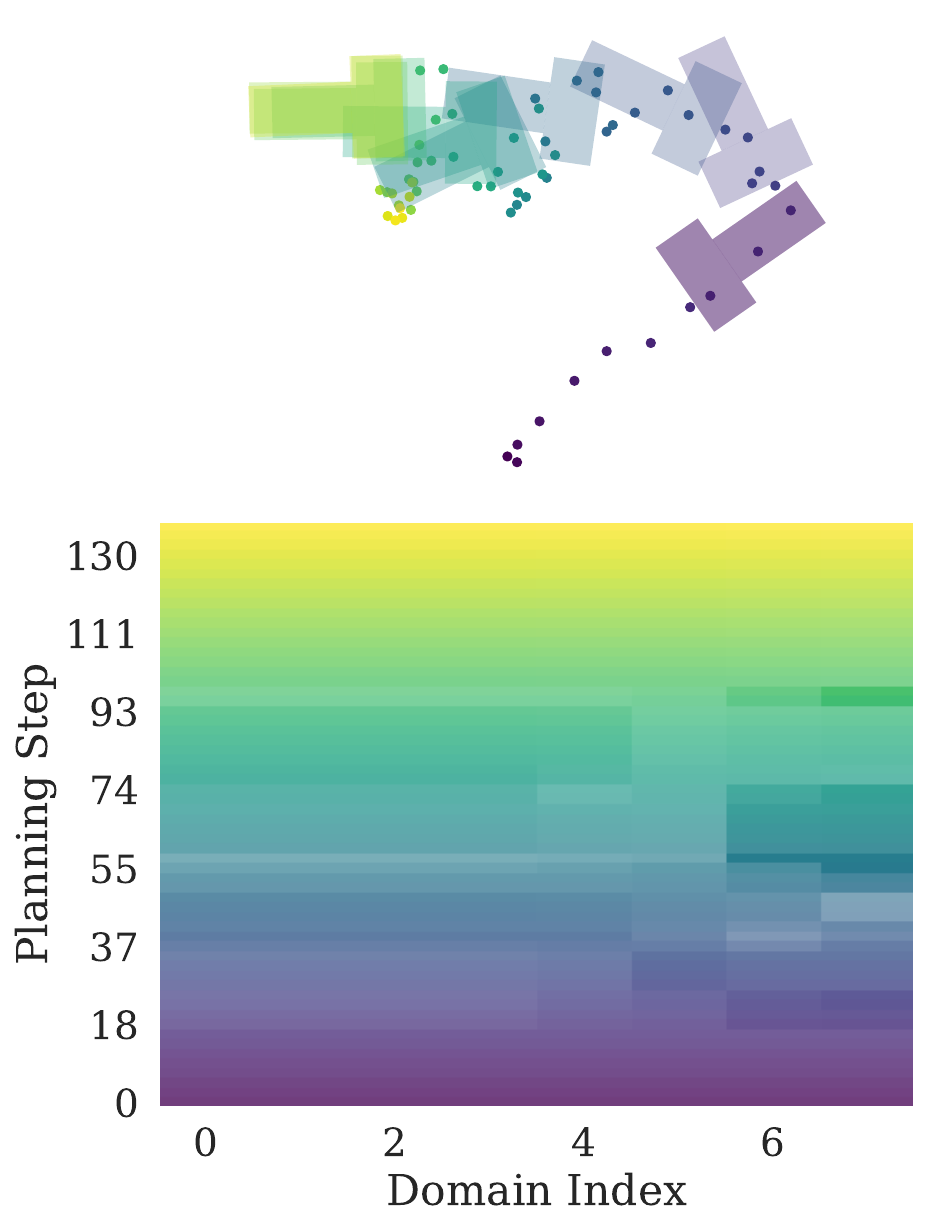}
  \caption{Exemplary evolution of domain weights for the domain-randomized Push-T task under margin randomization. Time is encoded by color and weight magnitude by transparency. Margins are randomized uniformly over $8$ domains within $\pm 1\,\si{cm}$ around the end-effector.}
  \label{fig:dr-margin-weights}\
  \vspace{-0.8cm}
\end{figure}

\begin{figure}[t]
    \centering
    \includegraphics[width=0.3\textwidth]{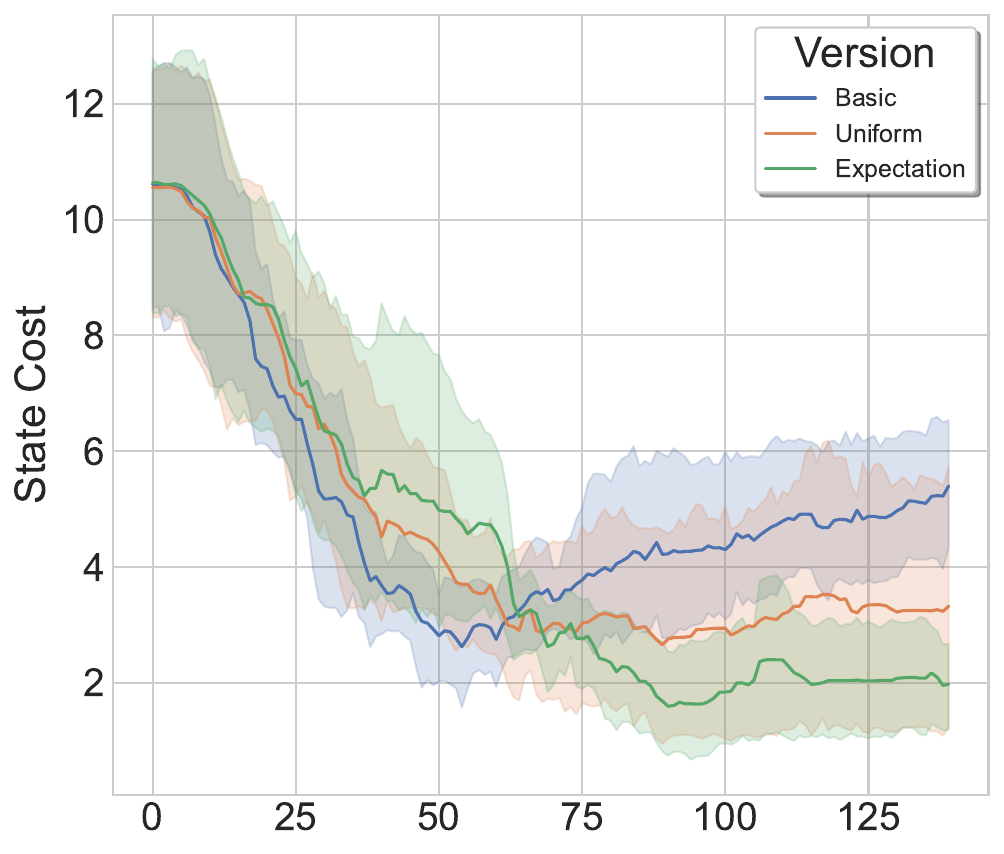}
    \caption{Real-world results of the MPC cost over time steps for different domain-randomization strategies.}
    \label{fig:dr-robustness}
    \vspace{-0.5cm}
\end{figure}

Domain randomization splits the sample budget across $D$ parallel domains.
We sweep over global physics parameters (friction, mass) and contact-initiation parameters (end-effector collision margins) across $6$ diverse seeds with $8$ domains, $160$ samples per domain, planning at $5~\si{\hertz}$.
 
\textbf{Mass and friction} yield no stable adaptation signal. Physics is dominated by contact timing and constraint resolution forces rather than by smooth variation in the randomized parameter. The resulting error signal is asymmetric: low-mass/friction domains overshoot and accumulate large pose errors, while high-mass domains move less and can appear spuriously consistent with the observation — causing adaptive weights to drift toward heavy/high friction domains.

\textbf{Contact margins} produce interpretable, mode-dependent weight evolution (\Cref{fig:dr-margin-weights}): weights shift toward larger margins during turning and away during straight pushing, consistent with margins directly gating contact onset. On hardware, kinematic misalignment causes the non-randomized baseline to accumulate error after initial pose correction; both uniform DR and adaptive reweighting improve final pose quality (\Cref{fig:dr-robustness}), though the advantage of adaptive over uniform weighting is small and not yet statistically conclusive.

\section{Conclusion}
\label{sec:conclusion}
We presented a high-fidelity MuJoCo MJX-based SMPC framework and systematically evaluated in simulation and on real hardware. 
The MTP algorithm consistently outperforms unimodal baselines in the Bugtrap Escape task, since its mixed global-local sampling is the only strategy that reliably escapes the local minimum.
On Push-T, it achieves the lowest final pose error and best variance in both simulation and real-robot trials.

Real-time deployment exposes a fundamental tension: small timesteps are required to avoid deep interpenetrations, yet shorter horizons degrade receding-horizon performance on tasks where long-range credit assignment matters.

Additionally, global physics parameters (mass, friction) are effectively non-identifiable within a single replanning interval, whereas contact-initiation parameters (collision margins) directly gate contact onset, yielding interpretable adaptation signals and measurable robustness gains. This is a structural limitation of the online SMPC setting that offline RL-based DR~\cite{tobin2017domain} does not share, since training amortizes contact noise across thousands of episodes.

As future work, to mitigate the longer-horizon problem, we will explore integrating learned value-function surrogates and generative sampling priors.

\section*{Acknowledgements}
This work is supported by ERC grant SIREN (101163933). Funded by the European Union. Views and opinions expressed are however those of the author(s) only and do not necessarily reflect those of the European Union or the European Research Council Executive Agency. Neither the European Union nor the granting authority can be held responsible for them.

\bibliographystyle{IEEEtran}
\bibliography{literature/lit, literature/thesis} 
\end{document}